\documentclass[dvipsnames]{article}

\usepackage[utf8]{inputenc}
\usepackage[T1]{fontenc}
\usepackage[USenglish]{babel}
\usepackage{csquotes}

\usepackage{amsmath}
\usepackage{amsfonts}
\usepackage{color}
\usepackage{placeins}

\usepackage{hyperref}
\usepackage{cleveref}
\usepackage{graphicx}
\usepackage{layouts}

\usepackage[a4paper, total={5.8in, 8.5in}, footskip=40pt, footnotesep=20pt]{geometry}
\usepackage{authblk}

\makeatletter
\renewcommand\AB@authnote[1]{\rlap{\textsuperscript{\normalfont#1}}}

\makeatother

\usepackage{fancyhdr}
\usepackage[ttscale=.85]{libertine}
\usepackage{libertinust1math}
\usepackage[activate={true,nocompatibility}, final, tracking=true, kerning=true, spacing=true, factor=1100, stretch=10, shrink=10]{microtype}
\microtypecontext{spacing=nonfrench}

\SetExtraSpacing{ encoding = {TS1} } { }

\usepackage{fnpct} % Minor improvements in footnote placement + kerning

\usepackage[
backend=bibtex,
style=numeric-comp,
maxcitenames=2,
natbib
]
{biblatex}

\emergencystretch=\maxdimen
\usepackage{xcolor}
\hypersetup{
    colorlinks,
    linkcolor={red!50!black},
    citecolor={blue!50!black},
    urlcolor={blue!80!black}
}

\usepackage{tikz}
\usetikzlibrary{positioning}
\usetikzlibrary{arrows}

\begin{document}

\title{Are demographically invariant models and representations in medical imaging fair?}

\author[1, 2, *]{Eike Petersen}
\author[3]{Enzo Ferrante}
\author[2, 4, 5]{Melanie Ganz}
\author[1, 2]{Aasa Feragen}
\affil[1]{Technical University of Denmark, DTU Compute}
\affil[2]{Pioneer Centre for AI, Denmark}
\affil[3]{Research Institute for Signals, Systems and Computational Intelligence (CONICET), Universidad Nacional del Litoral, Argentina}
\affil[4]{University of Copenhagen, Department of Computer Science, Denmark}
\affil[5]{Neurobiology Research Unit, Rigshospitalet, Copenhagen, Denmark}
\affil[*]{Corresponding author. Contact: \href{mailto:ewipe@dtu.dk}{ewipe@dtu.dk}.}
\date{}

\maketitle

\begin{abstract}
%%%
Medical imaging models have been shown to encode information about patient demographics such as age, race, and sex in their latent representation, raising concerns about their potential for discrimination.
Here, we ask whether requiring models \emph{not} to encode demographic attributes is desirable.
%We consider different types of invariance with respect to demographic attributes -- marginal and class-conditional representation invariance, and counterfactual model invariance -- and lay out their equivalence to standard notions of algorithmic fairness.
We point out that marginal and class-conditional representation invariance imply the standard group fairness notions of demographic parity and equalized odds, respectively.
In addition, however, they require matching the risk distributions, thus potentially equalizing away important group differences.
%In particular, representation invariance requires matching the predicted risk distributions of different groups, thus potentially equalizing away important group differences.
Enforcing the traditional fairness notions directly instead does not entail these strong constraints.
Moreover, representationally invariant models may still take demographic attributes into account for deriving predictions, implying unequal treatment -- in fact, achieving representation invariance may \emph{require} doing so.
In theory, this can be prevented using \emph{counterfactual} notions of (individual) fairness or invariance.
We caution, however, that properly defining medical image counterfactuals with respect to demographic attributes is fraught with challenges.
Finally, we posit that encoding demographic attributes may even be \emph{advantageous} if it enables learning a task-specific encoding of demographic features that does not rely on social constructs such as `race' and `gender.'
We conclude that demographically invariant representations are neither necessary nor sufficient for fairness in medical imaging.
Models may need to encode demographic attributes, lending further urgency to calls for comprehensive model fairness assessments in terms of predictive performance across diverse patient groups.
%\eikemargin{Abstract should be below 200 words}
%%%%
\end{abstract}

%\begin{titlepage}
%\maketitle
%\end{titlepage}

%\author{Eike Petersen \and Enzo Ferrante \and Melanie Ganz \and Aasa Feragen}
%\date{}
%\maketitle

%\begin{keywords}
%List of keywords, comma separated.
%\end{keywords}

\section{Introduction}
Recent studies have highlighted the capability of deep neural networks to infer patient demographics from medical images~\cite{Yi2021,Glocker2023,Gichoya2022,Duffy2022}.
In a widely cited study, \citet{Gichoya2022} found that neural networks can be trained to recognize a patient's race from chest X-ray recordings with high accuracy, even when medical doctors cannot, and even when correcting for confounding factors that correlate with racial identity.
Prompted by this work, \citet{Glocker2023} observed that even networks trained for chest X-ray-based disease classification (and not for inferring patient demographics) tend to learn a latent image encoding that differs in distribution between demographic groups.
Both studies warn that the ability and apparent tendency of such networks to encode patient identity enables -- intentional or unintentional -- explicit discrimination against demographic groups: if the latent representation encodes group membership, the model can exploit this in its decision-making.
For example, the model might learn to exploit spurious correlations between demographic groups and outcomes in the training dataset, which do not hold on another (real-world) dataset.
These cautionary notes are timely and consequential, and they prompt the question: can and \emph{should} we develop models that do \emph{not} encode group membership in their latent space?

\begin{figure*}[p]
    \centering
    \includegraphics[width=\textwidth]{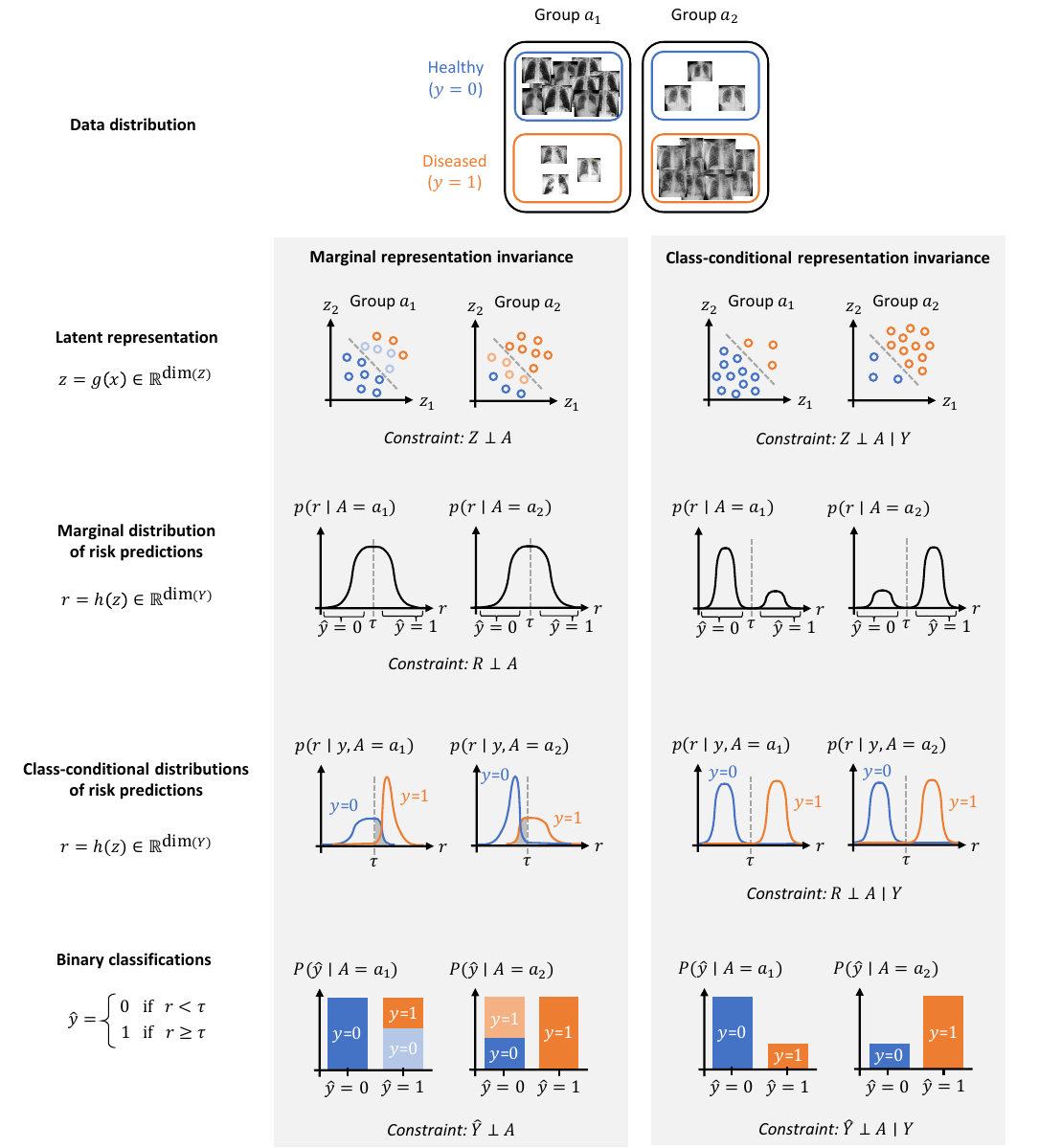}
    \caption{An illustration of the effects of enforcing marginal and class-conditional representation invariance in the case of a disease distribution with prevalence differences between two groups ($a_1$ and $a_2$). 
    The shown example is an illustrative case in which perfect classification is possible.
    Representation invariance implies risk distribution and classification rate invariances due to the deterministic relationships between the latent representation, risk predictions, and binary classifications.    
    Note the enforced misclassification (shaded) in the case of marginal invariance, and the enforced identity across groups of the marginal and class-conditional predicted risk distributions, respectively.
    While class-conditional representation invariance solves the illustrated issue due to prevalence differences, it still has drawbacks: it is incompatible with group-wise model calibration (see \cref{sec:conditional-invariance}) and requires equalizing away potentially important differences between unknown disease subtypes (see \cref{sec:general-drawbacks} and \cref{fig:intra-class}).
    }
    \label{fig:overview}
\end{figure*}

Here, we add to this ongoing discussion by considering different types of demographic invariance and assessing their relationship to notions of algorithmic fairness, as well as their desirability in the medical imaging domain.
First, we consider \emph{marginal representation invariance}, corresponding to a latent space that does \emph{not} encode group membership (\cref{sec:marginal-invariance}).
We point out that marginal invariance implies \emph{statistical parity}, and we discuss why this is highly undesirable in all cases in which disease prevalences differ between groups.
Secondly, we consider \emph{class-conditional representation invariance}, which implies the \emph{separation} fairness criterion and, as a consequence, equalized error rates (\cref{sec:conditional-invariance}).
We lay out why enforcing marginal or conditional invariance may be undesirable even in cases in which disease prevalences are correctly accounted for, since both notions of invariance require equalizing the predicted risk distributions across patient groups (\cref{sec:general-drawbacks}).
Notably, directly enforcing statistical parity or separation does \emph{not} require such equalization of risk prediction distributions.
\Cref{fig:overview} shows an illustration of the two representation invariance approaches.
Finally, we consider \emph{counterfactual model invariance}, corresponding to \emph{counterfactual fairness} (\cref{sec:model-invariance}).
We discuss the significant challenges involved in defining meaningful counterfactuals with respect to demographic groups in medical imaging, finding that this notion appears impractical in many cases and reduces to the two invariance notions discussed above in others, entailing the discussed drawbacks.
Our findings lend further urgency to the development and implementation of thorough fairness assessments and more nuanced unfairness mitigation techniques, given that models \emph{will} likely have the capability to discriminate against demographic groups.

The technical results we discuss are (mostly) not new; references to relevant prior work are provided throughout.
Our aim here is to emphasize the relevance of these results to the current discourse in (fair) medical imaging, and to explore their practical implications. % identify / emphasize / highlight / point out / spell out

%Here, we aim to add to this ongoing conversation by shedding light on one of the \emph{reasons} for a network to ``encode'' group membership in its latent representation.
%In particular, we point out that optimal prediction \emph{requires} encoding group membership in the latent representation $p(z)$ of the input data~$x$ if the label distribution $p(y\mid a)$ differs between patient groups~$a$.
%(Here, $x$ might denote blood test results, images, or various physiological measurements, and $y$ could represent, for example, disease labels, segmentation masks, or, in the case of generative models, images.)
%One important situation in which $p(y\mid a)$ differs between groups is the case of group-dependent disease prevalence.
%We furthermore show why a demographically invariant representation is often undesirable \emph{even in the case of identical label distributions~$p(y \mid a)$}.
%We conclude that a latent representation that does \emph{not} encode group membership appears undesirable in almost all practical applications.

%Following this first observation, we furthermore explore the applicability and ethical desirability of other forms of demographic invariances in medical imaging, specifically 

\subsection{Problem setting and notation}
We consider a supervised prediction setting in which the objective is to predict labels~$Y$ from inputs~$X$.
Patients are assumed to be members of different and potentially overlapping demographic groups~$A$ (e.g., young, female, or young female).
The inputs~$X$ might denote medical images, blood markers, or other physiological measurements, and the labels~$Y$ could, e.g., represent categorical disease labels, continuous disease severity scores, or segmentation masks.
We will often consider a latent representation~$Z$ of the input data~$X$, obtained using a deterministic feature extractor~$g\colon x \mapsto z$, such as the first layers of a deep neural network.
This latent representation is then mapped onto predictions~$r\in\mathbb{R}^{\mathrm{dim}(Y)}$ by a second deterministic model component~$h\colon z \mapsto r$.
Depending on the application, $r$ may then be discretized using a decision threshold~$\tau$ to obtain binary outcome labels~$\hat{y}$ representing, for example, disease labels or segmentation masks.
Neither~$g$ nor~$h$ receive group membership as an explicit input, and we assume~$\tau$ to be group-independent.
We will refer to random variables by upper-case letters ($X$, $Y$, $Z$) and to their realizations by lower-case letters ($x$, $y$, $z$).
%\new{Refer to \cref{sec:group-dependency} for a critical appraisal of these assumptions and potential alternatives.}
We consider a model to `encode' group membership~$A$ in its latent representation~$Z$ if group membership can be inferred with higher-than-chance accuracy from~$Z$.

\section{Marginal representation invariance}
\label{sec:marginal-invariance}
%\section*{Optimal models must encode group membership if $p(y \mid a)$ differs between groups}
Aiming for a model that does \emph{not} encode group membership is equivalent to requiring the latent representation~$z$ to be invariant between groups~$a$, that is,
\begin{equation}
    p(z) = p(z\mid a) \quad \forall \, a \quad \Leftrightarrow \quad Z \perp A.
    \label{eq:marginal-alignment}
\end{equation}
%In other words, finding a model that does not encode group membership is equivalent to finding a model that has a demographically invariant latent representation.
%Otherwise, group membership could be predicted with higher-than-chance accuracy from the latent representation.
%However, from \cref{eq:marginal-alignment}, 
However, if~$p(z\mid a)$ is identical across groups, then the distributions of model predictions~$p(r \mid a)$ and~$p(\hat{y} \mid a)$ are identical across groups as well, implying $R, \hat{Y} \perp A$.
This corresponds to the algorithmic fairness notion of \emph{statistical parity}, or \emph{independence}~\cite{Zemel2013,Zhao2022,Barocas2019}.
If, however, the true outcome distributions~$p(y\mid a)$ differ between groups, we cannot require \cref{eq:marginal-alignment} to hold without incurring a loss in predictive accuracy since $p(\hat{y} \mid a) \neq p(y \mid a)$ necessarily~\cite{Dwork2012,Hardt2016a,ricci2022addressing}.

\begin{figure}[t]
    \centering
    \includegraphics{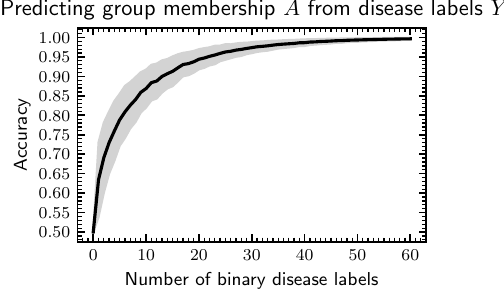}
    \caption{
    Accuracy with which binary group membership~$A$ can be predicted just from the target label~$Y$, \emph{fully ignoring any input data~$X$}, as a function of the dimensionality of~$Y$. The label~$Y$ is assumed binary-valued here, representing, e.g., multiple binary disease labels.
    Incidences per disease (i.e., elements of $Y$) are drawn randomly from~$[0.1, 0.9]$ for the two groups; mean and standard deviation of the resulting prediction accuracy over 1000 repetitions shown.
    (In the case of identical prevalences, a label does not increase the identifiability of~$A$.)
    Note that if~$A$ is identifiable from~$Y$, the same is true for any accurate model predictions~$\hat{Y}$.
    Illustrates why asking for groups to be non-identifiable in the presence of prevalence differences is ill-advised, and how this problem is aggravated for higher-dimensional~$Y$, such as in multi-class settings or segmentation.}
    \label{fig:y-ident}
\end{figure}

Why would~$p(y \mid a)$ differ between groups in a medical application?
In the simplest case, \emph{prevalence} could differ: A classifier that returns the same distribution of positive and negative predictions for breast cancer in men and women appears undesirable as the prevalence of breast cancer differs between these groups. 
If, on the other hand, the classifier returns positive and negative predictions at different rates for the different groups (as would be correct), then its latent representation is statistically related to group membership, thus `encoding' group membership.
%However, differences in~$p(y \mid a)$ may not arise only from prevalence differences.
As another example from medical image segmentation, organ shapes or sizes may vary depending on age or sex, resulting in differences in the distribution~$p(y \mid a)$ of true segmentation masks between groups~\cite{Peters2006}.
%In this case, one should, again, expect the latent space of an image-level segmentation algorithm (such as the one described by \citet{Larrazabal2020}) to encode group membership.\melmargin{But they are doing classification aka diagnosing 14 common thoracic diseases using X-ray images, not segmentation, right? -- Eike: Ahh sorry, this was supposed to be a reference to Enzo's image-level segmentation paper. Enzo, can you add that? -- Enzo: Not sure to what paper you are referring Eike. Is it our TMI paper where we show that segmentation models for X-ray images fail in young and old patients ( https://ieeexplore.ieee.org/abstract/document/9963582 )? or the one where we analyze the latent space for different medical centers ( https://arxiv.org/pdf/2211.07395.pdf )? -- Aasa: Would it be possible to give just a little more explanation of what is going wrong in whichever paper you are referring to? That would make the discussion easier to follow}
%The same is true for medical image generation models, in which the label distribution is a distribution over images.
Notice that this issue becomes worse for high-dimensional~$y$ as it becomes progressively easier to identify~$a$ based on distributional differences between~$p(y \mid a)$ for the different groups, see \cref{fig:y-ident} for an illustration of this effect.

In domain adaptation theory, property~\eqref{eq:marginal-alignment} is called \emph{marginal alignment}, and a latent representation that satisfies it is often simply called \emph{domain-invariant}.
Extensive research has been devoted to domain-invariant representation learning, and by identifying different \emph{groups} with different \emph{domains}, these results apply to our setting~\cite{Mukherjee2022a}.
The equality of $p(y\mid a)$ between groups is known to be a necessary condition for the existence of a predictive performance-optimal (marginally) invariant representation~\cite{Li2018c,Zhao2022}.
Similarly, the field of algorithmic fairness has widely discussed the inherent tensions between statistical parity and predictive accuracy.
Enforcing statistical parity despite differences in~$p(y \mid a)$ is known to incur a significant loss in predictive performance~\cite{Dwork2012,Barocas2019,McNamara2019,Zhao2022}.
For binary classification with two sensitive groups, %the loss in accuracy resulting from the use of a demographically invariant representation has been characterized by 
\citet{Zhao2022} show that enforcing statistical parity implies that the sum of the error rates in both groups is lower-bounded by the prevalence difference. % $\lvert p(y \mid a_1) - p(y \mid a_2) \rvert$.
%classification accuracy will decrease by at least half of the base rate difference $\lvert p(y \mid a_1) - p(y \mid a_2) \rvert$.\footnote{This is a slight simplification of Theorem 3.1 by \citet{Zhao2022}, considering only the case of equally-sized groups~$a_1$ and~$a_2$.}

To provide some intuition concerning the theoretical results presented in this section, note that there is a crucial difference between a \emph{model} (i.e., a function) and a \emph{representation} being invariant with respect to a demographic feature. %~$A$ and, on the other hand, the \emph{representation}~$Z$ being (marginally) invariant with respect to that feature.
The former asserts that the prediction~$\hat{y}$ should be the same regardless of the value of~$A$ while keeping everything else fixed; we will explore this notion in more detail in \cref{sec:model-invariance}.
Importantly, this still allows for the predictions~$\hat{Y}$ and group membership~$A$ to be correlated if~$A$ correlates with disease-relevant features.
A (marginally) invariant \emph{representation}, on the other hand, requires full statistical independence~$Z,\hat{Y} \perp A$, even if the labels~$Y$ are correlated with~$A$.
In this case, a demographically invariant representation %not only \emph{ignores} disease-relevant information such as prevalence differences but also 
requires \emph{actively confounding} predictions to comply with the invariance requirement, enforcing unequal treatment and unnecessary misclassification in all groups.
(Note that in the case of differing $p(y \mid a)$, a perfect classifier that predicts all labels correctly does not satisfy marginal independence.)
Maybe counter-intuitively, enforcing statistical independence will also often result in disparate treatment of subjects from different groups~\cite{Lohaus2022}: 
if the distribution of disease presentations differs between groups, but the model is forced to map them to an identical latent distribution, then this can only be achieved by treating similar subjects from different groups dissimilarly.
(In this case, the unequal treatment occurs in the model's feature extraction -- or \emph{encoding} -- step.)
%The performance loss incurred by requiring marginal independence also surpasses the frequently observed `leveling down' phenomenon~\cite{Zhang2022, Zietlow2022, Zhao2022}, since, in the case of differences in $p(y \mid a)$ between groups, marginal independence is fundamentally incompatible with performance-optimal prediction~\cite{Dwork2012, Hardt2016a, Zhao2022}.
%(In this case, a perfect classifier does not satisfy marginal independence.)

In the case of low-dimensional labels such as disease labels, there is one obvious strategy for addressing the challenges posed by label shift (of which prevalence differences are one instance) for marginal invariance approaches: 
One can artificially balance the training set such that $p(y \mid a)$ is equal across groups.
In some cases, one may also consider prevalence differences between groups negligible or purely an artifact of biased data collection procedures.
Would enforcing a demographically invariant representation appear beneficial in such cases?
%And, seeing that it addressed the label shift problem, does class-conditional representation invariance generally appear desirable?
We will return to this question in \cref{sec:general-drawbacks}, where we discuss more general drawbacks of representation invariance approaches.

%Firstly, notice that the property~\ref{eq:marginal-alignment} is inherently tied to the dataset it is evaluated on: once one moves from an artificially balanced dataset to a real-world dataset in which the labels are no longer equally distributed across groups, the latent representation will no longer be demographically invariant, and groups can, again, be identified with some accuracy from the model's latent encoding of its inputs.
%Secondly, this approach would still suffer from the issues related to active equalization of within-class differences between groups that were discussed in the previous paragraph.
%Thirdly, groups can still be treated arbitrarily differently...

\section{Class-conditional representation invariance}
\label{sec:conditional-invariance}
The situation in which $p(y \mid a)$ differs between domains is known as \emph{label shift}~\cite{TachetdesCombes2020,Li2020c}, and various methods have been developed to address it.
One class of proposed solutions~\cite{TachetdesCombes2020,Li2018c} attempts to match the \emph{class-conditional} distributions~${p(z \mid y)}$ across groups, instead of the \emph{marginal} distributions~$p(z)$ like in the previous section.
Formally, this requirement can be stated as 
\begin{equation}
    p(z \mid y) = p(z\mid y, a) \quad \forall \, a, y \quad \Leftrightarrow \quad Z \perp A \mid Y.
    \label{eq:class-conditional-invariance}
\end{equation}
Analogously to the previous section, observe that \cref{eq:class-conditional-invariance} implies equality of~$p(r \mid y, a)$ and~$p(\hat{y} \mid y, a)$ across groups, and thus $R, \hat{Y} \perp A \mid Y$.
This implies the \emph{separation} criterion in algorithmic fairness~\cite{Barocas2019}. %\footnote{Note that separation does \emph{not} imply class-conditional representation invariance.}
Note that class-conditional representation invariance does \emph{not} imply marginal representation invariance, i.e., models that satisfy \cref{eq:class-conditional-invariance} may encode group membership in their latent space.

Is class-conditional representation invariance a desirable property for medical imaging models? %, even leaving aside the question of the desirability of encoding group membership for a moment?
We argue here that this is not the case, even leaving aside the fact that group membership may be encoded.
%Firstly, consider the simple case of a binary classification task, i.e., $y, \hat{y} \in \{0, 1\}$, and a probabilistic classifier that maps the latent representation~$z$ to risk scores~$r\in[0, 1]$, which are then turned into binary predictions~$\hat{y}$ by considering some group-independent decision threshold~$\tau$.
%In this case, \cref{eq:class-conditional-invariance} also implies $R \perp A \mid Y$, i.e., separation in terms of the model's risk predictions.
It is known that a risk score~$R$ cannot satisfy separation and be \emph{well-calibrated by groups} at the same time, except in pathological cases~\cite{Barocas2019}. %\footnote{It is important to distinguish between separation in terms of risk scores~$R$ and separation in terms of categorical model predictions~$\hat{Y}$; refer to~\citet{Reich2020, Petersen2023} for details.}
For this reason, and because its intrinsic ethical significance appears unclear, it has been argued that separation \emph{at the risk score level} (${R \perp A \mid Y}$) may not be a desirable model property~\cite{Reich2020,Hedden2021,Petersen2023}.
Notably, this is distinct from separation \emph{at the classification level} (${\hat{Y} \perp A \mid Y}$), which entails equal error rates between groups and may be a desirable aim.
However, as has been pointed out elsewhere~\cite{Reich2020,Petersen2023}, error rate equality can be achieved without the overly restrictive requirement of class-conditionally invariant risk distributions.
Further fundamental drawbacks of both marginal and class-conditional representation invariance approaches are discussed in the following section.
%Additional fundamental drawbacks of both marginal and class-conditional representation invariance will be discussed in the following section.

\section{General drawbacks of representation invariance}
\label{sec:general-drawbacks}
\begin{figure*}[t]
    \centering
    \includegraphics[width=\textwidth]{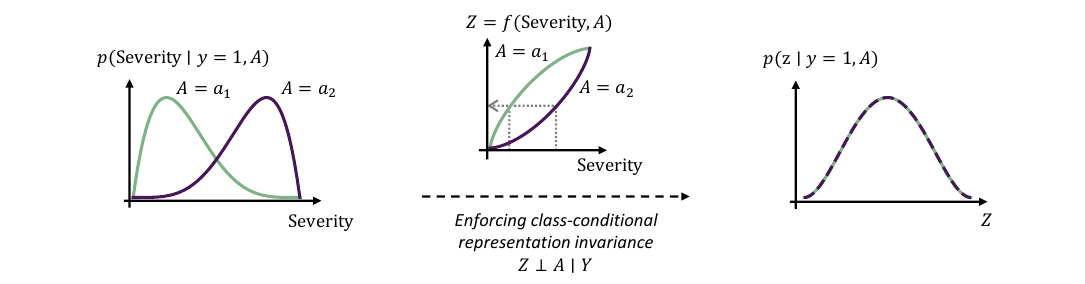}
    \caption{An illustration of the effects of enforcing (class-conditional) representation invariance in the case of differences in the within-class distributions between two groups ($a_1$ and~$a_2$). 
    The distribution of the (unlabeled/unobserved) disease severity may differ between sick patients of the two groups (left panel).
    Enforcing (class-conditionally) identical latent representations~$Z$ (right panel) requires mapping patients with differing disease severities to the same latent representation, depending on which group a patient belongs to (center).
    This will also result in them being assigned the same risk predictions, see \cref{sec:conditional-invariance}.
    }
    \label{fig:intra-class}
\end{figure*}

A key drawback of any notion of representation invariance concerns their neglect of differences in the \emph{within-class distributions} between groups~\cite{Li2020c,Mahajan2021}.
A single class label often subsumes many fundamentally distinct situations, such as different subtypes or presentations of the disease, a phenomenon known as \emph{hidden stratification}~\cite{OakdenRayner2020a}.
The situation in which disease manifestations differ between groups (or domains) has also been called \emph{manifestation shift}~\cite{Castro2020}.
If the distribution of label manifestations differs between groups, then representation invariance can only be achieved by forcing distinct stratifications in different groups to be represented by the same latent encoding.
This is analogous to the higher-level effects of enforcing marginal invariance in the case of label shift, discussed in \cref{sec:marginal-invariance}.
One instance of the within-class variation problem %that is also not addressed by matching approaches 
concerns differences in task difficulty~\cite{Petersen2023a}.
For physiological or technical reasons, a prediction task may be harder in one group than in others, due to, for example, differences in the amount of (breast, fat) tissue that confounds a recording~\cite{Alexander1958,Brahee2013}.
In this case, a predictive model should return less confident predictions for members of the more strongly confounded group, even for the same disease instance.
Representation invariance approaches require equalizing such essential differences between groups (see \cref{fig:intra-class} for an illustration).
This example also illustrates again why representation invariance does not guarantee `fair treatment,' and, in fact, may \emph{require} disparate treatment~\cite{Lohaus2022}.
Different images may be mapped to an identical or similar latent representation in order to achieve representation invariance; an argument for why this may actually be \emph{enforced} by distribution matching approaches was provided above.

Do representation invariance approaches provide any meaningful fairness guarantees, then?
As we have seen, marginal invariance implies statistical parity, which may be a desirable goal in scarce resource distribution settings or patient prioritization~\cite{Barocas2019,Petersen2023} but is typically not considered desirable in predictive settings.
In this regard, \citet{Glocker2023} also point out that if a representation \emph{encodes} group membership, this does not imply that the model also \emph{uses} this information for decision-making.
Conditional invariance, on the other hand, implies equalized error rates, which may be a desirable aim in the predictive setting, even though its algorithmic enforcement is known to suffer from the leveling down phenomenon~\cite{Zhang2022,Zietlow2022}.
However, equalized error rates can also be achieved \emph{without} requiring equalized conditional risk distributions~\cite{Reich2020,Petersen2023}.
Thus, the conditional representation invariance requirement may be seen as overly restrictive, resulting in the noted incompatibility with calibration by groups.

A further limitation of the fairness guarantees provided by representation invariance approaches concerns their inherent dependency on the dataset on which they are evaluated.
Representation invariance is a property of a model \emph{applied to a particular dataset}, and there is no guarantee that the same model will produce an invariant representation (and, say, equal error rates) on another dataset, due to potential (label, covariate, intra-class variation) distribution shifts.
Representation invariance thus provides no guarantee about a particular (fairness-relevant) property of the model itself.
This is especially concerning in cases in which (as is general best practice) training and evaluation datasets are carefully curated and thus may differ importantly from real-world observational data.
(Note that these limitations apply not only to the discussed representation invariance approaches, but also to all other statistical group fairness criteria.)

To address the drawbacks of representation matching approaches, various alternative approaches have been proposed in the domain adaptation literature, including object matching~\cite{Li2020c,Mahajan2021}, relaxed distribution matching~\cite{Johansson2019, Wu2019a}, support alignment~\cite{Tong2022}, and hypothesis invariance~\cite{Wang2021c}.
A convincingly successful application of such approaches to the medical imaging domain remains to be demonstrated, however.

%\section{Is representation invariance the right aim?}
At this point, let us step back and ask: why were we interested in demographically invariant representations in the first place?
%Returning to the study of \citet{Gichoya2022}, the authors note that the capability of models to predict racial identity provides ``a direct vector for the reproduction or exacerbation of the racial disparities that already exist in medical practice.''
%They further caution that this process may lead the model to make race-specific errors, which would be difficult to detect for clinicians.
Echoing similar concerns raised by \citet{Gichoya2022}, \citet{Glocker2023} write that ``an algorithm may encode protected characteristics, and then use this information for making predictions due to undesirable correlations in the (historical) training data.''
These remarks reflect a concern that patients with similar (patho)physiology might wrongly receive different predictions if the model takes their demographic attributes into account.
We argue that this concern is more aligned with requiring demographically invariant \emph{models}, as opposed to demographically invariant \emph{distributions}.
We will thus consider model invariance approaches next.

\section{Model invariance}
\label{sec:model-invariance}
To address the discussed drawbacks of invariant representation learning, there has been a shift in the domain generalization community away from learning invariant \emph{representations} toward learning invariant \emph{mechanisms}, that is, models or model components that are, in some sense, invariant with respect to a nuisance variable~$A$~\cite{Arjovsky2019,Scholkopf2021,Subbaswamy2019,Puli2022,Makar2022}.
One popular approach is based on the notion of \emph{counterfactual invariance}: `keeping everything else fixed' (we will elaborate on what this may mean below), changing just the attribute~$A$ should not change model predictions~\cite{Veitch2021}.
When~$A$ represents a sensitive attribute, this corresponds to \emph{counterfactual fairness}~\cite{Kusner2017}, a variant of individual (as opposed to group) fairness~\cite{Dwork2012}.
In theory, counterfactual invariance holds across datasets: if a \emph{model} is invariant with respect to changes in~$A$, then this will still hold if the data distribution changes.
%We thus argue that \emph{model} invariance, not \emph{representation} invariance should be the property of interest for fairness considerations.
%The case of invariance with respect to demographic groups in medical imaging is particularly complex, however, and challenging to define formally.
While such theoretical guarantees can indeed be obtained for invariances with respect to formally defined algebraic operations on the input space such as rotations or translations~\cite{Bekkers2018, Aslan2023}, the case of invariance with respect to demographic groups in medical imaging is more complex, however, and challenging to define formally.

\begin{figure}[t]
    \centering
    \tikzset{
        %Define standard arrow tip
        >=stealth',
        %Define style for boxes
        punkt/.style={
               rectangle,
               rounded corners,
               draw=black, very thick,
               text width=6.5em,
               minimum height=3em,
               text centered},
        % Define arrow style
        pil/.style={
               ->,
               thick,
               shorten <=2pt,
               shorten >=2pt,},
    }
    \small{
    \begin{tikzpicture}[node distance=1.5cm, auto,]
         %nodes
         \node[punkt] (sex) {Birth Sex};
         \node[punkt, below=0.8cm of sex] (bio) {Anatomy \& Physiology};
         \node[punkt, right=1cm of sex] (gender) {Gender};
         \node[punkt, below=0.8cm of bio] (x) {$x$};
         \node[punkt, below=0.8cm of gender] (lf) {Lifestyle Factors};
         \node[punkt, below=0.8cm of lf] (disease) {Disease};
        
         %edges
         \draw[pil] (sex) -- (bio);
         \draw[pil] (bio) -- (x);
         \draw[pil] (sex) -- (gender);
         \draw[pil] (gender) -- (lf);
         \draw[pil] (lf) -- (disease);
         \draw[pil] (disease) -- (x);
         \draw[pil] (bio) -- (disease);
         \draw[pil] (lf) -- (bio);
         \draw[pil] (bio) -- (lf);
    \end{tikzpicture}}
    \caption{A simplistic causal diagram illustrating some of the many ways in which biological (birth) sex may causally influence medical image recordings~$x$. 
    With respect to which of the causal paths from ``Birth Sex'' to~$x$ should a disease classification model be (counterfactually) invariant?}
    \label{fig:causality}
\end{figure}

Discussions of counterfactual fairness and counterfactual invariance are predominantly framed in terms of \emph{causal models}~\cite{Pearl2009, Kusner2017, Chiappa2019, Veitch2021, Scholkopf2021}, and both causal modeling generally and counterfactual inference in particular have recently received increased attention in the medical imaging community~\cite{Castro2020, Ribeiro2023, Pawlowski2020, Ouyang2023, Vlontzos2022}.
Consider the highly simplistic causal diagram~\cite{Pearl2009} in \cref{fig:causality}, representing the diverse effects of biological birth sex in a hypothetical medical imaging scenario.
With respect to which of the causal paths from ``Birth Sex'' to~$x$ should a disease classification model be (counterfactually) invariant?
Invariance with respect to all of them would correspond to fully equalizing the (disease-conditioned) distributions of image representations of different biological (birth) sexes.
In a chest X-ray context, this might imply adjusting female torso shapes, tissue distributions, organ sizes, and differing disease presentations to conform with the distribution observed in the male group.
Adjusting all of these factors when creating, for example, a `male counterfactual' of a female image, yields a fully `male-typical' example.
It is thus not surprising, that, for example, \citet{Veitch2021} found such a counterfactual invariance approach to 
imply $R, \hat{Y} \perp A \mid Y$, i.e., \emph{separation} -- just like in the case of class-conditional representation invariance, thus entailing the same drawbacks (also cf.~\citet{Nilforoshan2022}).
\citet{Puli2022} derive similar equivalences.
Note that enforcing such an invariance, just like the invariant representation approaches discussed above, would also change how mere correlates of biological sex such as body mass index (BMI) are permitted to influence model predictions.

On the other hand, maybe we would like the model to only be invariant with respect to \emph{some} of the paths in \cref{fig:causality}, resulting in \emph{path-specific} counterfactual fairness~\cite{Chiappa2019}?
While certainly feasible, the authors are unaware of any works implementing such path-specific invariances in the context of medical image analysis.
This may be owed to the complexity of identifying the different causal relationships in a causal diagram such as \cref{fig:causality} from empirical data.
While progress has been made toward this goal~\cite{Pawlowski2020}, such approaches typically require observations of all variables in the causal graph, which are often unavailable.
Moreover, any such approach crucially depends on the correctness and completeness of the specified causal graph.

As we have seen, the crux of counterfactual invariance approaches thus %, and the reason for the results of \citet{Veitch2021} and \citet{Puli2022}, 
lies in the precise definition of the counterfactuals to consider.
What, precisely, would we like to keep fixed, and what should be adjusted, when generating `biological sex counterfactuals', `ethnicity counterfactuals', or `age counterfactuals'?
Which are the peculiar properties of group~$a$ with respect to which a model should be invariant, and which are mere correlates that we would permit to influence model decisions~\cite{Chiappa2019, Kasirzadeh2021}?
These are highly nontrivial questions, particularly in the context of \emph{social constructs} such as demographic groups~\cite{Kasirzadeh2021}, and in particular as it relates to medical imaging. %that may not admit a satisfactory answer.
The validity of a counterfactual model invariance approach fully depends on the answers, however.
As a final note of caution, any causal conception of fairness is crucially vulnerable to unobserved confounding~\cite{Kilbertus2020}.

%Also recall again that a demographically invariant representation is a property of a model \emph{given a particular dataset} and will likely no longer hold on another, potentially more realistic, dataset that follows a different distribution.
%Both of these are not true in the case of an invariant \emph{model}: 

%\section{Benefits of encoding group membership}
%Given the outlined drawbacks of \emph{preventing} the encoding of group membership in a model's latent space, we ask:

\section{Discussion \& Conclusion}
Our starting point in this piece were observations by different research groups that deep learning models tend to encode patients' demographic features in their latent space~\cite{Glocker2023,Gichoya2022}.
Prompted by these observations, we first asked whether it is feasible and desirable to train models that do \emph{not} encode demographic group membership, in the sense that it cannot be inferred with higher-than-chance accuracy from the latent representation.\footnote{While we adopt this terminology here, we note that it may be considered misleading to write that a latent representation \emph{encodes} group membership simply because that representation is \emph{correlated} with group membership. The representation may in fact encode only disease-relevant features, but if those features are correlated with group membership, this will also be true for the latent representation -- even if no group-specific features are encoded.}
Noting that this requirement is equivalent to \emph{marginal representation invariance} and implies \emph{statistical parity}, we discussed why such a requirement can severely hurt model performance~\cite{Dwork2012,Barocas2019,Zhao2022,McNamara2019,Zhao2022a} even beyond the well-known `leveling down' effect~\cite{Zhang2022,Zietlow2022}.
%At the same time, we discussed how marginal representation invariance does not guarantee fair treatment of patients from different demographic groups.
We then considered a close alternative to marginal representation invariance, \emph{class-conditional invariance}~\cite{TachetdesCombes2020}.
This notion alleviates one of the problems associated with marginal invariance: severe model performance deterioration in the face of label shift, i.e., differences in~$p(y\mid a)$ between groups.
However, class-conditional invariance turned out to imply the \emph{separation} criterion from algorithmic fairness (on the risk score level), which may be considered undesirable due to its incompatibility with group-wise calibration as well as for other reasons~\cite{Reich2020,Hedden2021,Petersen2023,Barocas2019,Mehta2024}.
Directly enforcing error rate equality across groups instead does not entail the same drawbacks.
Moreover, all representation invariance approaches suffer from their neglect of differences in \emph{intra-class variations} between groups due to, for example, different disease subtypes or manifestations~\cite{OakdenRayner2020a,Li2020c,Mahajan2021,Castro2020}.
In addition, they do not guarantee `fair treatment,' and, in fact, may even \emph{require} disparate treatment~\cite{Lohaus2022} -- which is simply moved to the model's feature extraction stage.
Finally, we considered counterfactually invariant \emph{models} as one potential alternative~\cite{Veitch2021,Puli2022,Kusner2017,Chiappa2019}, finding that both the theoretical definition and practical inference of meaningful demographic counterfactuals in the medical imaging domain is fraught with serious challenges and often incurs the same drawbacks as representation invariance approaches~\cite{Nilforoshan2022}.

%\subsubsection*{Can it be \emph{beneficial} for a model to encode group membership?}
%\textbf{Can it be \emph{beneficial} for a model to encode group membership?}
Can it even be \emph{beneficial} for a model to encode group membership?
Consider the case of a model that depends on explicit group membership information.
The model might exploit such information during training -- e.g., to achieve representation invariance -- or during inference, in order to perform group-specific calibration or thresholding.
%to make up for its invariance, for example by means of re-calibration as discussed above.
Such a model inherits the well-documented problems that arise from sorting humans into categories based predominantly on \emph{sociocultural} constructs~\cite{Tomasev2021,Belitz2022,Andrews2023}, as opposed to physiological differences relevant to the prediction task at hand.
As an alternative, consider now the case of a model that has learned an \emph{implicit} representation of group membership.
This second model can learn to extract just those aspects of, say, ethnicity or biological sex that are relevant to the given prediction task.
Moreover, it can learn a \emph{continuous} representation of these properties, as opposed to coarse, human-defined categories.
This might enable more robust generalization to patient groups that fall between traditional categories, such as nonbinary or mixed-race patients.
To be clear: this is, of course, a highly idealized view.
Whether a model indeed learns such a minimal and optimal representation in practice is an important question.
Encouragingly, \citet{Glocker2023} provide evidence that this may indeed occur in practice, at least in some instances.
They find that a model trained only for disease prediction, and not demographic group prediction, indeed appears to learn a feature representation that is far less predictive of demographic properties than the representation learned by a model trained explicitly for demographic group prediction.

%\subsubsection*{Representation invariance and generalization.}
%\textbf{Representation invariance and generalization.}
While our discussion of representation invariance has been largely negative, we want to emphasize that such approaches are not without merit.
Their main field of application is the unsupervised domain adaptation setting, in which labels~$y$ are unavailable from a new domain.
In this case, invariant representation learning can be beneficial~\cite{Li2018c,Moyer2020}.
Notably, this differs from the setting we consider here, in which labels are available from all demographic groups.
Nevertheless, such invariant representations might generalize better to previously unseen demographic groups, even though domain generalization benchmarks appear to refute this intuition~\cite{Wang2022c}. %Gulrajani
Moreover, our discussion here concerned the case of enforcing \emph{strict} invariance.
In practice, many invariant representation learning approaches instead use regularization to prevent learning representations that are \emph{unnecessarily strongly} predictive of the domain or demographic group~\cite{Zhao2022,Brown2023}.
Such approaches may prove beneficial if they are tuned to minimize the negative effects of enforcing strict invariance.
Similar benefits may be obtained using less restrictive mitigation techniques such as fair pruning~\cite{Wu2022} or fair early exiting~\cite{Chiu2023}.
Using such methods, the latent representation will still encode demographic group membership to some degree, however.

%\subsubsection*{Implications of our analyses.}
%\textbf{Implications of our analyses.}
%We have pointed out here that if there are label differences between groups, an optimal classifier will necessarily encode group membership in its latent space.This has important consequences for model fairness assessments. 
Taken together, our analyses imply that, at present, there is no straightforward method for developing medical imaging models that guarantee similar treatment of individuals with similar physiology but differing demographic features.
Enforcing different notions of demographic invariance is likely to harm different aspects of model performance across all groups, as has been observed empirically~\cite{Zietlow2022,Zhang2022,Mehta2024}, while at the same time not providing any meaningful guarantees in terms of equal treatment.
As a consequence, we claim that the simple fact that a model encodes group membership cannot be understood as a fairness violation.
We have argued that group encoding can in some cases even be considered beneficial if it prevents models from inheriting the biases encoded in human-defined patient categories.
Our analysis lends further urgency to calls for fairness assessments in terms of comprehensive subgroup analyses since, as pointed out before~\cite{Glocker2023,Gichoya2022}, models that encode demographics \emph{have} the capacity to discriminate.
For improving model performance in under-performing groups, simple balanced empirical risk minimization, targeted data collection efforts, and investigations into the potential presence of label biases currently appear to be among the most promising practical avenues~\cite{Zhang2022,Idrissi2022,Petersen2023a}.
If the aim is to prevent demographic shortcut learning, methods such as the one proposed by~\citet{Brown2023} could be used instead of demographic invariance.%, which does not provide any guarantees in this regard.

As a final remark, we note that while our discussion here was centered around medical image analysis, our analysis and discussion applies readily to other medical and non-medical application domains.
%Moreover, as we have emphasized, \aasamarkup{model developers should still strive to minimize the degree to which models encode such patient characteristics to the minimum degree necessary.}{There is one other aspect here that you are not mentioning. If you try to remove demographic information through e.g. adversarial training or an additional component in your loss function, then you make your model more complex. In part, this makes it even harder to decipher what is going on in the model. In part, it might also make the loss function harder to optimize. And in part, you risk making the model less stable. I don't think we are ready to claim that removing such information is necessarily desirable.}

\subsubsection*{Author contributions}
All authors contributed to the conceptualization of the manuscript.
EP wrote the original draft, and all authors contributed to manuscript reviewing and editing.
EF, MG and AF were responsible for funding acquisition, and MG and AF for supervision.

\subsubsection*{Funding acknowledgements}
Work on this project was partially funded by the Independent Research Fund Denmark (DFF, grant number 9131-00097B), Denmark's Pioneer Centre for AI (DNRF grant number~P1), a Google Award for inclusion research, the Novo Nordisk Foundation through the Center for Basic Machine Learning Research in Life Science (MLLS, grant number NNF20OC0062606), and the Argentinian National Agency for the Promotion of Research, Technological Development and Innovation (ANPCyT, grant number PICT-PRH-2019-0000).
The listed funding sources had no involvement in the conception of the manuscript, the writing of the report, or the decision to submit the article for publication.
%\uselengthunit{mm}\printlength{\textwidth}

%\subsection*{Impact statement}
%Our work addresses the implications of imposing certain technical constraints on machine learning models for the purpose of improving model fairness.
%As we point out, such constraints are not without drawbacks and their application should be considered carefully, especially in healthcare.
%While our arguments here are theoretical and academic in nature, work such as ours might affect future regulatory healthcare efforts and thus have a tangible impact on, e.g., the quality of future patient care (for better or worse).
%Algorithmic fairness is a complex and multi-faceted concept and objective that requires careful deliberation and stakeholder involvement; rash implementation of simplistic fairness conceptions may do more harm than good.

\printbibliography

\end{document}